\documentclass{article}

\usepackage{algorithm}
\usepackage{algpseudocode}
\usepackage{amsmath,amsfonts,amssymb}
\usepackage{graphicx}
\usepackage{setspace}
\usepackage{tocloft}

\usepackage{PRIMEarxiv}

\usepackage[utf8]{inputenc} 
\usepackage[T1]{fontenc}    
\usepackage{hyperref}       
\usepackage{url}            
\usepackage{booktabs}       
\usepackage{amsfonts}       
\usepackage{nicefrac}       
\usepackage{microtype}      
\usepackage{lipsum}
\usepackage{fancyhdr}       
\usepackage{graphicx}       
\graphicspath{{media/}}     

\title{Three-dimensional Tracking of a Large Number of High Dynamic Objects from Multiple Views using Current Statistical Model
}

\author{
  Nianhao Xie \\
  College of Aerospace Science and Engineering \\
  National University of Defense Technology \\
  Changsha, China \\
  \texttt{{xienianhao}@nudt.edu.cn} \\
}

\begin{document}
\maketitle

\begin{abstract}
Three-dimensional tracking of multiple objects from multiple views has a wide range of applications, especially in the study of bio-cluster behavior which requires precise trajectories of research objects. However, there are significant temporal-spatial association uncertainties when the objects are similar to each other, frequently maneuver, and cluster in large numbers. Aiming at such a multi-view multi-object 3D tracking scenario, a current statistical model based Kalman particle filter (CSKPF) method is proposed following the Bayesian tracking-while-reconstruction framework. The CSKPF algorithm predicts the objects' states and estimates the objects' state covariance by the current statistical model to importance particle sampling efficiency, and suppresses the measurement noise by the Kalman filter. The simulation experiments prove that the CSKPF method can improve the tracking integrity, continuity, and precision compared with the existing constant velocity based particle filter (CVPF) method. The real experiment on fruitfly clusters also confirms the effectiveness of the CSKPF method.
\end{abstract}


\section{Introduction}\label{sec:intro}

Multi-object tracking (MOT) are widely applied on traffic control, sports analysis, battlefield reconnaissance, etc. In the tracking of a biol-cluster, e.g. fish school, birds flock and insect swarm \cite{attanasiCollectiveBehaviourCollective2014,watzek2021Modelling,shelton2020Collective} , there are more objects and higher motion kinematics compared with the traditional MOT scenario, which makes the general MOT methods not applicable. Moreover, 3D trajectories of biol-cluster are needed which requires images from synchronized multiple cameras to do object 3D reconstruction at the same time.
    
\par The 3D multi-view multi-object tracking faces two major challenges compared with the single-object tracking \cite{ThreedimensionalObjectTracking2018} or static objects 3D reconstruction \cite{shiReconstructionDenseThreedimensional2013} .  One is temporal association problem, i.e. how to determine the detection\footnote{For the sake of clarity, the concepts of \emph{measurement}, \emph{detection}, and \emph{obervation} in this paper are firstly be claimed. On an image, the foreground pixels of objects form many \emph{measurements}. The measurement of a certain object is its \emph{detection} and the state reconstructed by the detection is its \emph{observation}.} association of each object between the current frame and the previous frames on a certain view. The other one is the spatial association problem, i.e. how to determine the detection association of each object on different views in a certain frame. Muti-view multi-object tracking is essentially a problem of assigning temporal-spatial association of detections. The temporal association decides the 2D tracking while the spatial association decides the 3D reconstruction. 

\par For the temporal association problem, the associated detection on the current frame to the previous frame can be constrained to a smaller region using motion continuity constraints. For the spatial association problem, the associated detection on one view to another view can be constrained on the epipolar. Assuming that the association credibility between two measurements can be defined, there are statistical-based methods that assign the nearest measurement to the target detection, such as joint probability data association filter (JPDAF) and multiple hypothesis tracking (MHT), which is effective when the number of objects is not large and the measurement distinction is not small \cite{coxEfficientImplementationEvaluation1994} . The methods based on linear allocation establish a cost matrix between the target detection and the candidate measurement, and then the Hungarian algorithm is used to solve the optimal detection-measurement allocation problem\cite{ duRelativeEpipolarMotion2007} .
    
\par The order of reconstruction and tracking is important. According to the relationship between the temporal association and spatial association, the existing multi-view multi-object tracking method can be divided into three categories: a) Firstly, temporal associations are assigned to generate 2D trajectories on each view, and then the spatial association of these 2D trajectories is assigned to reconstruct 3D trajectories, named tracking-to-reconstruction method \cite{wuAutomated3DTrajectory2011} . This type of algorithm cannot effectively use the 3D motion prediction of the object to deal with the occlusion problem, which often leads to trajectory interruptions and switch. b) Firstly, the spatial association is assigned to reconstruct the 3D points in each frame, and then the temporal association of these 3D points is assigned to link 3D trajectories, named reconstruction-to-tracking method \cite{ardekaniThreedimensionalTrackingBehaviour2013} . This type of method often results in many illusory 3D trajectories when the measurement is not discriminative enough, because the spatial association will generate many "ghost" 3D points. c) Assuming the initial 3D position of each object is known, firstly, the 3D position in the next frame is predicted by the kinematic model and reprojected to each view by spatial association constraints. The prediction can be confirmed if there is overlap between the reprojected detection and the measurements on every view, which makes the temporal association natural satisfies the spatial association constraint. This type of method, named tracking-while-reconstruction method \cite{chengNovelMethodTracking2015} , can effectively eliminate phantoms and is insensitive to occlusion, which can obtain continuous and reliable trajectories.
    
\par In many cases, especially for bio-clusters, it is difficult to obtain an object's kinematic model. Chen et al. \cite{chengNovelMethodTracking2015} used the assumption of uniform linear motion to predict the object's position which is reasonable when the frame rate is large enough. Wang et al. \cite{wangTracking3DPosition2017} used the LSTM network to learn the kinematic model of a single fruitfly, which makes the position prediction more reasonable but largely increases calculation burden. Considering that the individual's motion is a variable acceleration motion with no sudden acceleration, in this paper, the current statistical model (CSM) \cite{zhouJiDongMuBiaoGenZong1991} is used to predict not only the position but also the velocity of the object. We establish a tracking-while-reconstructing framework based on Bayesian inference and solve the temporal-spatial association problem by particle filter method, where the object's states and covariance are predicted by the current statistical model. After the detections are assigned to each object, the state observations are determined. The Kalman filter (KF) is used to suppress measurement noise and improve tracking precision. The proposed method, named current statistical model based Kalman particle filter (CSKPF) tracking algorithm, can not only build the trajectories with higher integrity, continuity, and precision but also can estimate the target velocity during tracking, which is essential to analyze the clusters behaviors.
    
\par This paper is arranged as follows: Section \ref{sec:intro} introduces the significance and challenges of multi-view multi-object tracking, especially the 3D tracking of a large number of objects with high kinematics, and summarizes existing temporal-spatial association methods and tracking-reconstruction frameworks. Section \ref{sec:method} extends the Bayesian inference framework from single-view single-object tracking to the multi-view multi-object tracking-while-reconstructing. Then, the existing constant velocity based particle filter (CVPF) method is explain in the above framework, which leads to our CSKPF method. Section \ref{sec:test} first defines the evaluation index for multi-view multi-object tracking, and then the tracking performance of CVPF and CSKPF methods are compared by simulated and real data, which proves the improvement of the proposed method. Finally, the paper is summarized in section \ref{sec:conclusion}.
    
\section{Method}\label{sec:method}
In this section, the Bayesian framework for single-view single-object tracking is reviewed and the reason why it cannot be directly applied to multi-view multi-object tracking is analyzed. Then, a multi-view multi-object Bayesian inference framework to deal with the above problems is built. Finally, the existing CVPF method and the proposed CSKPF method based on this framework are explained.
    
\subsection{The Bayesian single-view single-object tracking framework}
 The discrete kinematic model and obervation model of the target can be formulated as
\begin{equation}
    \mathbf{x}_{t} = \mathbf{f}\left( \mathbf{x}_{t-1},\mathbf{v}_{t-1} \right),
    \mathbf{y}_{t} = \mathbf{h}\left( \mathbf{x}_t,\mathbf{n}_t \right),
\end{equation}
where $\mathbf{x}_t, \mathbf{y}_t$ are the state and obervation of the target at moment $t$, and $\mathbf{v}_t, \mathbf{n}_t$ are the process noise and obervation noise, and $\mathbf{f}$ and $\mathbf{h}$ are the state transition function and measurement function, respectively.

\par From the Bayesian point of view, the object tracking is to estimate the current state of $\mathbf{x}_t$ based on historical obervation $\mathbf{y}_{1:t}$. Assuming that the state transition is a first-order Markov process, and the \emph{posterior} state probability $p\left(\mathbf{x}_{t-1}|\mathbf{y}_{1: t-1} \right)$ at time $t-1$ is known, then the \emph{priori} state at time $t$ can be \emph{predicted} by the state transition model $p\left( \mathbf{x}_t|\mathbf{x}_{t-1}\right)$ before getting the obervation $\mathbf{y}_t$ at moment $t$, i.e.
\begin{equation}\label{equ:bayes1}
    p\left( \mathbf{x}_t|\mathbf{y}_{1:t-1}\right)=\int p\left( \mathbf{x}_t|\mathbf{x}_{t-1}\right)
    p \left(\mathbf{x}_{t-1}|\mathbf{y}_{1:t-1} \right) \,d\mathbf{x}_{t-1} .
\end{equation}

After the obervation $\mathbf{y}_t$ is known, the \emph{priori} state can be \emph{corrected} to \emph{posterior} state

\begin{equation}\label{equ:bayes2}
    p\left( \mathbf{x}_t|\mathbf{y}_{1:t}\right) \propto  p\left( \mathbf{y}_t|\mathbf{x}_t\right) p\left( \mathbf{x}_t|\mathbf{y}_{1:t-1}\right),
    \end{equation}
where $p\left( \mathbf{y}_t|\mathbf{x}_t\right)$ means the probability that an obervation $\mathbf{y}_t$ occurs when the state is $\mathbf{x}_t$.

\par It seems that by generating as many trackers as there are objects and making each tracker run independently, the multi-view multi-object tracking problem can be solved by the above single-view single-object framework. However, on the one hand, the target detection cannot be determined before the temporal association of detection; on the other hand, the target observation cannot be reconstructed before the spatial association of detection. In brief, $\mathbf{y}_t$ is unknown because the temporal-spatial association is pendent, which causes the \emph{correct} step of Bayesian inference unimplementable. An improved Bayesian framework is designed in next subsection to determine the temporal-spatial association which makes the single-object Bayesian implementable in a multi-view multi-object tracking scenario. 

\subsection{A Bayesian multi-view multi-object tracking-while-reconstruction framework}\label{sec:frame}

\begin{figure}[!t]
    \centering
    \includegraphics[width=12cm]{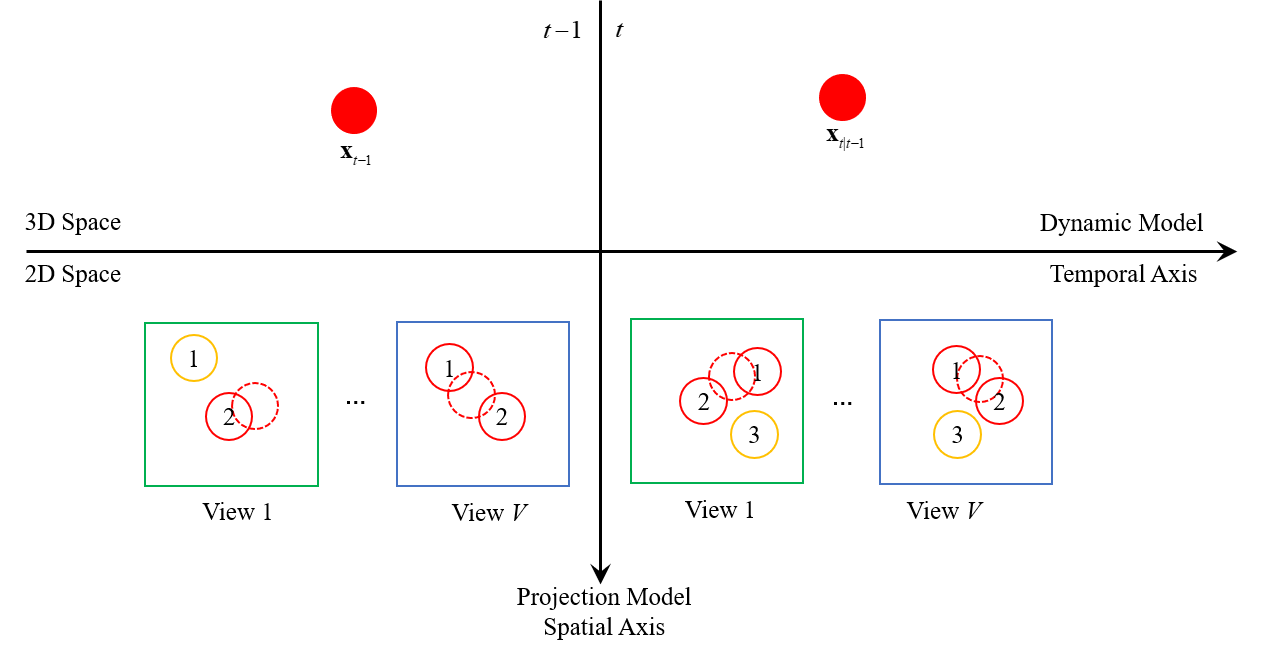}
    \caption{A Bayesian multi-view multi-object tracking-while-reconstruction framework. The filled circles are 3D state and the dashed circles are their projection on each view. The solid circles are the 2D measurements, which are colored red/yellow if they are overlapping/non-overlapping with the state projection.}\label{fig:frame}
\end{figure}

As mention above, the framework treat each object independently. Without loss of generality, as shown in Figure~\ref{fig:frame}, only one tracker's temporal-spatial association Bayesian inference will be explained. The horizontal axis is the temporal axis driven by the kinematic model, and the left half-plane represents the previous moment $t-1$, and the right half-plane is the current moment $t$. The vertical axis is the spatial axis determined by the camera projection model, the upper half-plane is 3D space, and the lower half-plane represents the 2D space of all views. Supposed that the state $\mathbf{x}_{t-1}$ has been tracked according to the historical measurements. Then the \emph{prior} state $\mathbf{x}_{t|t-1}$ can be predicted based on the kinematic model $p\left(\mathbf{x}_t|\mathbf{x}_{ t-1}\right)$. Finally, the projection of $\mathbf{x}_{t|t-1}$ on each view can be calculated by the camera projection model. The state transfers from the third quadrant through the second and first quadrants to the fourth quadrant is constrained by the temporal association constraints and spatial association constraints. 
    
\par In the fourth quadrant, the temporal-spatial association can be determined by the position and appearence similarity between the target and candidate detection, which means the target observation $\mathbf{y}_t$ is known. To be specific, in the fourth quadrant, if the projection $\mathbf{P}^v(\mathbf{x}_{t|t-1})$ has overlap with measurement $\chi _{v,t,k_v}$ on view $v$, then the candidate $\mathbf{z}_{t,k}=\langle \chi _{1,t,k_{1}},\cdots,\chi _{V,t,k_{V}} \rangle, k=1,\cdots, K$ is a potential temporal-spatial association. All potential candidates constitute the \emph{candidate association group} $\mathbf{z}_t=\{\mathbf{z}_{t,k}\}_{k=1}^K$. The credibility of the candidate association is defined as
    \begin{equation}\label{equ:weights} 
        w(\mathbf{z}_{t,k})=p\left(\mathbf{z}_{t,k}|\mathbf{x}_{t|t-1}\right)=\prod_{v=1}^V  {\frac{1}{\left(\sqrt{2\pi}\sigma_v\right)^{-\frac{V}{2}}} \exp \left({\tau_v \left(\mathbf{x}_{t|t-1}, \chi_{v,t,k_v} \right)-1}\right)},
    \end{equation}
    where $\tau_v \left(\mathbf{x}_{t|t-1}, \chi_{v, t,k_v} \right)$ is the appearence similarity between the projection $\mathbf{P}^v(\mathbf{x}_{t|t-1})$ and measurement $\chi_{v,t,k_v}$. Thus, the target observation is the most reliable candidate association
    \begin{equation}\label{equ:maxRelGroup}
    \mathbf{y}_t = \mathop{\arg \max} \limits_{k}{w(\mathbf{z}_{t,k})}.
    \end{equation}
    According to the Bayesian inference formula (\ref{equ:bayes2}), the corrected target state is
    \begin{equation}\label{equ:bayes3}
    \mathbf{x}_t=\mathop{\arg \max} \limits_{\hat{\mathbf{x}}_t}{p\left( \hat{\mathbf{x}}_t|\mathbf{y} _{1:t}\right)}.
    \end{equation}
    
    
    \par The proposed framework can track muliple objects while reconstruction. The tracker for each object runs independently in parallel and one measurement can be associated with more than one object, which improves tracking efficiency and avoids tracking interruption caused by image occlusion in some views.
    
    \subsection{The constant velocity based particle filter (CVPF) method}\label{sec:cvpf}
    It is diffcult to solve the above tracking framework because there is an integral term in Equation~(\ref{equ:bayes1}). Therefore, the Monte Carlo sampling method is used to solve the Bayesian inference problem \cite{candyBayesianSignalProcessing2016}, i.e. particle filter method. Specifically, Chen et al. \cite{chengNovelMethodTracking2015} proposed the CVPF method (as shown in Algorithm~\ref{algo:CVPF}) based on constant velocity kinematic model and sampling importance resampling (SIR).
    
    \begin{algorithm}
        \footnotesize
        \caption{CVPF algorithm} \label{algo:CVPF}
        \hspace*{0.02in} {\bf Parameter:} number of particles $N_p$, particle variance $\sigma^2$\\
        \hspace*{0.02in} {\bf Input:} the measurements $\mathcal{B}_{t,v}$ of each frame $t$ and each view $v$\\
        \hspace*{0.02in} {\bf Output:} 3D trajectories of all objects\\
        \hspace*{0.02in} {\bf Initialization:} \\
        \hspace*{0.28in} Calculate temporal-temporal association among $\mathcal{B}_{1,v}, \mathcal{B}_{2,v}, v=1,\cdots, V$ using method of exhaustion and obtain $N_2$ matching pairs. All associated measurements on the second frame are put into the associated measurement set $\mathcal{M}_{2,v}$. Create $N_2$ new trackers and put them into the active tracker set $\mathcal{T}_2$ and initialize the inactive tracker set $\mathcal{L}_2$ as empty set.\\ 
        \begin{algorithmic}[1]
        \For{$t$ = $3,\cdots,T$}
        \State Initialize $\mathcal{M}_{t,v}=\emptyset$
            \For {$\tau$ in $\mathcal{T}_{t-1}$}
            \State \emph{Predict} the state $\mathbf{x}_{t|t-1}$ of the tracker $\tau$ from $\mathbf{x}_{t-1}$ using Equation~(\ref{equ:kcf1}).
            \State \emph{Sample} $N_p$ particles $\mathbf{x}^i_{t|t-1} \sim \mathcal{N}\left(\mathbf{x}_{t|t-1}, \sigma^2 \right)$ and renew their weights by $1/N_p$.
                \For {$i=1,\cdots,N_p$}
                    \State Calculate $\mathbf{z}^i_{t|t-1}$ and their credibilities for $\mathbf{x}_{t|t-1}^i$ using Equation~(\ref{equ:weights}).
                    \State Determine the particle weights $w_i$ using Equation~(\ref{equ:weightsCal}).
                \EndFor
                \If {$\forall i, w_i=0$}
                    \State Move the tracker $\tau$ from the active tracker set $\mathcal{T}_{t-1}$ to inactive $\mathcal{L}_{t-1}$.
                \Else
                    \State \emph{Estimate} the state $\hat{\mathbf{x}}_{t}$ by Equation~(\ref{equ:particleSum}).
                    \State Calculate $\mathbf{z}_t$ its credibilities for $\hat{\mathbf{x}}_{t}$ using Equation~(\ref{equ:weights}).
                    \State Put the most likely association $\mathbf{z}_t^*$ into the association measurements set $\mathcal{M}_{t,v}$.
                    \State Reconstruct the most likely obervation $\mathbf{y}_t$ by $\mathbf{z}_t^*$ using stereovision method.
                    \State $\mathbf{x}_{t} \leftarrow \hat{\mathbf{x}}_{t}$.
                \EndIf
            \EndFor
            \State Do temporal-spatial association among the unassociated measurements $\mathcal{B}_{t-1,v}/\mathcal{M}_{t-1,v}$, $\mathcal{B}_{t,v }/\mathcal{M}_{t,v}$ using method of exhaustion and initialize new trackers $\mathcal{R}_{t}$.
            \State Update the tracker set by $\mathcal{T}_{t} \leftarrow \mathcal{T}_{t-1} \bigcup  \mathcal{R}_{t}$.
        \EndFor
        \State Parse the trajectories of all objects from the historical states of trackers $\mathcal{T}_T$ and $\mathcal{L}_T$.
        \end{algorithmic}
    \end{algorithm}
    
    \subsubsection{State prediction}
    \par Denote the target state as $\mathbf{x}=[x,\dot{x},y,\dot{y},z,\dot{z}]^{\top}$, the precited state based on constant velocity hypothesis is
    \begin{equation}\label{equ:kcf1}
        \mathbf{x}_{t|t-1}=\mathbf{F}\mathbf{x}_t,
    \end{equation}
    where $\mathbf{F}$ is the state transition matrix
    \begin{equation}
        {\bf{F}} = \left[ {\begin{array}{*{20}{c}}
            {{{\bf{F}}_1}}&{\bf{0}}&{\bf{0}}\\
            {\bf{0}}&{{{\bf{F}}_1}}&{\bf{0}}\\
            {\bf{0}}&{\bf{0}}&{{{\bf{F}}_1}}
            \end{array}} \right], ~\mathrm{and}~~\bf{F}_1=\left[ {\begin{array}{*{20}{c}}
            1&\Delta t\\
            0&1
            \end{array}} \right],
    \end{equation}
    and $\Delta t$ is the sampling interval.
    \subsubsection{Particle sampling}
    \par Suppose there is $N_p$ particles, the importance probability density function of the particle $\mathbf{x}_{t|t-1}^i,i=1,\cdots, N_p$, is $q\left( {{\bf{x}}_t^i|{\bf{x}}_{t-1}^i,{ {\bf{y}}_{1:t}}} \right) = p\left( {{\bf{x}}_t^i|{\bf{x}}_{t-1}^i } \right)$ according to the SIR filter method. The CVPF method assumes that $p\left( {{\bf{x}}_t^i|{\bf{x}}_{t-1}^i} \right)$ obeys Gaussian distribution with mean value $\mathbf{x}_{t|t-1}$ and variance $\sigma^2$. By sampling $N_p$ particles from the the importance probability density function $\mathbf{x}_{t|t-1}^i \sim \mathcal{N}(\mathbf{x}_{t|t-1},\sigma ^2)$, and renewing the particle weights as $w_i=1/N_p$, the candidate association group $\mathbf{z}_{t|t-1}^i$ of $i$-th particle and its credibility can be calculated according to the tracking framework. 
    
    \subsubsection{State correlation}
    According to Equation~(\ref{equ:weights}), the particle weights are updated to
    \begin{equation}\label{equ:weightsCal}
        w_i = \mathop{\max} \limits_{k}{w(\mathbf{z}_{{t|t-1},k}^i)},
    \end{equation}
    where $\mathbf{z}_{{t|t-1},k}^i$ is the $k$-th candidate association of the particle $\mathbf{x}_{t|t-1}^i$. Thus, the estimated target state is
    \begin{equation}\label{equ:particleSum}
        {\hat{\mathbf{x}}_{t}} = \sum\limits_{i = 1}^n {{w_i}{\bf{x}}_{t|t - 1}^i}.
    \end{equation}
    
    \par The CVPF method takes the ${\hat{\mathbf{x}}_{t}}$ as the state correction ${\mathbf{x}_{t}}$. By reprojecting the state ${\hat{\mathbf{x}}_{t}}$ to each view, the most likely association $\mathbf{z}_t^*$ can be determined according to the Equation~(\ref{equ:maxRelGroup}) and the observation $\mathbf{y}_t$ can be calculated by stereovision method. Finally, new trackers can be initialized from the unassociated measurement.
    
    \subsection{Currect statistical model based Kalman particle filter (CSKPF) method}
    The constant velocity hypothesis of CVPF method makes it be unsuitable for maneuvering target. Moreover, the particle covariance is always constant $\sigma^2$, which does not reflect the state prediction uncertainty. The proposed CSKPF method (as shown in Algorithm \ref{algo:CSKPF}) will improve the CVPF method by solving the two problems.
    \begin{algorithm}
        \footnotesize
        \caption{CSKPF algorithm} \label{algo:CSKPF}
        \hspace*{0.02in} {\bf Parameter:} number of particles $N_p$\\
        \hspace*{0.18in} observation noise matrix $\mathbf{R}$, initial process noise $\mathbf{Q}$, initial covariance $\mathbf{P}$\\
        \hspace*{0.18in} warm-up time $T_h$, reciprocal of maneuver time $\alpha_i$, maximum possible acceleration $a_{\mathrm{max},i}$\\
        \hspace*{0.02in} {\bf Input:} the measurements $\mathcal{B}_{t,v}$ of each frame $t$ and each view $v$\\
        \hspace*{0.02in} {\bf Output:} 3D trajectories of all targets\\
        \hspace*{0.02in} {\bf Initialization:} same as the CVPF method.
        \begin{algorithmic}[1]
        \For{$t$ = $3,\cdots,T_h$}
        \State $\divideontimes$  Execute CVPF and record the acceleration as the current average acceleration $\bar{\bf{a}}$.
        \EndFor
        \For{$t$ = $T_h+1,T_h+2,\cdots,T$}
            \State Initialize $\mathcal{M}_{t,v}=\emptyset$
            \For {$\tau$ in $\mathcal{T}_{t-1}$}
            \State * \emph{Predict} the state $\mathbf{x}_{t|t-1}$ of the tracker $\tau$  from $\mathbf{x}_{t-1}$ using the Equation~(\ref{equ:CSM1}).
            \State $\divideontimes$ \emph{Predict} the covariance $\mathbf{P}_{t|t-1}$ of the tracker $\tau$ using the Equation~(\ref{equ:CSM2}).
            \State * \emph{Sample} $N_p$ particles $\mathbf{x}_{t|t-1} \sim \mathcal{N}\left(\mathbf{x}_{t|t-1}, \mathbf{P}_{t|t-1} \right)$ and renew their weights by $1/N_p$.
                \For {$i = 1,\cdots, N_p$}
                    \State Calculate $\mathbf{z}^i_{t|t-1}$ and their credibilities for $\mathbf{x}_{t|t-1}^i$ using Equation~(\ref{equ:weights}).
                    \State Determine the particle weights $w_i$ using Equation~(\ref{equ:weightsCal}).
                \EndFor
                    \If {$\forall i, w_i=0$}
                    \State Move the tracker $\tau$ from the active tracker set $\mathcal{T}_{t-1}$ to inactive $\mathcal{L}_{t-1}$.
                \Else
                    \State \emph{Estimate} the state $\hat{\mathbf{x}}_{t}$ by Equation~(\ref{equ:particleSum}).
                    \State Calculate $\mathbf{z}_t$ its credibilities for $\hat{\mathbf{x}}_t$ using Equation~(\ref{equ:weights}).
                    \State Put the most likely association $\mathbf{z}_t^*$ into the association measurements set $\mathcal{M}_{t,v}$.
                    \State Reconstruct the most likely obervation $\mathbf{y}_t$ by $\mathbf{z}_t^*$ using stereovision method.
                    \State $\divideontimes$ \emph{Correct} the state $\mathbf{x}_{t}$ by Equation~(\ref{equ:kcf4}) and $\mathbf{P}_{t}$ by Equation~(\ref{equ:kcf5}).
                \EndIf
            \EndFor
            \State Do temporal-spatial association among the unassociated measurements $\mathcal{B}_{t-1,v}/\mathcal{M}_{t-1,v}$, $\mathcal{B}_{t,v }/\mathcal{M}_{t,v}$ using method of exhaustion and initialize new trackers $\mathcal{R}_{t}$.
            \State Update the tracker set by $\mathcal{T}_{t} \leftarrow \mathcal{T}_{t-1} \bigcup  \mathcal{R}_{t}$.
        \EndFor
        \State Parse the trajectories of all objects from the historical states of trackers $\mathcal{T}_T$ and $\mathcal{L}_T$.
        \end{algorithmic}
        {\bf Note:} * marks the revised steps, and $\divideontimes$ marks additional steps campared to CVPF method.
    \end{algorithm}
    
    \subsubsection{State prediction}
    \par The object motion is modeled by currect statistical model which assumes that the object acceleration is bounded and obeys modified Rayleigh distribution to the average acceleration in a past period of time. Denote the target state as $\mathbf{x}=[x,\dot{x},\ddot{x},y,\dot{y},\ddot{y},z,\dot{z}, \ddot{z}]^{\top}$ and predict the target state by
    \begin{equation}\label{equ:CSM1}
        \mathbf{x}_{t|t-1}=\mathbf{G}\mathbf{x}_t+\mathbf{U}\bar{\mathbf{a}},
    \end{equation}
    and estimate its covariance by
    \begin{equation}\label{equ:CSM2}
        \mathbf{P}_{t|t-1}=\mathbf{G} \mathbf{P}_{t-1} \mathbf{G}^{\top } + \mathbf{Q}_t, 
    \end{equation}
    where $\bar{\mathbf{a}} $ is the current average acceleration and
    \begin{equation}
        {\bf{G}} = \left[ {\begin{array}{*{20}{c}}
            {{{\bf{G}}_1}}&{\bf{0}}&{\bf{0}}\\
            {\bf{0}}&{{{\bf{G}}_1}}&{\bf{0}}\\
            {\bf{0}}&{\bf{0}}&{{{\bf{G}}_1}}
            \end{array}} \right], ~\mathrm{and}~\bf{G}_1=\left[ {\begin{array}{*{20}{c}}
            1&{\Delta t}&{\frac{1}{{{\alpha ^2}}}\left( { - 1 + \alpha \Delta t + {e^{ - \alpha \Delta t}}} \right)}\\
            0&1&{\frac{1}{\alpha }\left( {1 - {e^{ - \alpha \Delta t}}} \right)}\\
            0&0&{{e^{ - \alpha \Delta t}}}
            \end{array}} \right]
        \end{equation}
    is the state transition matrix, 
    \begin{equation}
        {\bf{U}} = \left[ {\begin{array}{*{20}{c}}
        {{{\bf{u}}_1}}&{\bf{0}}&{\bf{0}}\\
        {\bf{0}}&{{{\bf{u}}_1}}&{\bf{0}}\\
        {\bf{0}}&{\bf{0}}&{{{\bf{u}}_1}}
        \end{array}} \right], ~\mathrm{and}~{{\bf{u}}_1} = \left[ {\begin{array}{*{20}{c}}
        {\frac{1}{{{\alpha ^2}}}\left( { - \alpha \Delta t + \frac{{{\alpha ^2}{{\left( {\Delta t} \right)}^2}}}{2} + 1 - {e^{ - alpha \Delta t}}} \right)}\\
        {\frac{1}{\alpha }\left( {\alpha \Delta t - 1 + {e^{ - \alpha \Delta t}}} \right)}\\
        {1 - {e^{ - \alpha \Delta t}}}
        \end{array}} \right]
    \end{equation}
    is the input control matrix. The the process noise estimation matrix 
        \begin{equation}\label{equ:Qt}
            {\bf{Q}} = \left[ {\begin{array}{*{20}{l}}
                {{{\bf{Q}}_{1}}}&{\bf{0}}&{\bf{0}}\\
                {\bf{0}}&{{{\bf{Q}}_2}}&{\bf{0}}\\
                {\bf{0}}&{\bf{0}}&{{{\bf{Q}}_3}}
                \end{array}} \right], ~\mathrm{and}~
            \begin{array}{l}
                {{\bf{Q}}_i} = 2\alpha_i \sigma_i^2\left[ {\begin{array}{*{20}{c}}
                {{q_{11}(\alpha_i)}}&{{q_{12}(\alpha_i)}}&{{q_{13}(\alpha_i)}}\\
                {{q_{12}(\alpha_i)}}&{{q_{22}(\alpha_i)}}&{{q_{23}(\alpha_i)}}\\
                {{q_{13}(\alpha_i)}}&{{q_{23}(\alpha_i)}}&{{q_{33}(\alpha_i)}}
                \end{array}} \right],i=1,2,3
            \end{array}\\
        \end{equation}
    keeps changing over time, where 
    \begin{equation}
        \sigma _i^2 = \left\{ 
        {\begin{array}{*{20}{c}}
            {{\frac{4 - \pi}{\pi }}{{[{a_{{\rm{max}},i}} - \bar a_i(t)]}^2}{\rm{ }}}\\
            {{\frac{4 - \pi}{\pi }}{{[ - {a_{{\rm{max}},i}} - \bar a_i(t)]}^2}}
            \end{array}} \right.
                \begin{array}{*{20}{l}}
                {}&{\bar a_i(t) > 0}\\
                {}&{\bar a_i(t) < 0}
                \end{array},\\
    \end{equation}
    and 
    \begin{equation}
        \begin{array}{l}
        q_{11} = \frac{1}{{2{\alpha ^5}}}\left[ {1 - {e^{ - 2\alpha \Delta t}} + 2\alpha \Delta t + \frac{2}{3}{\alpha ^3}\Delta {t^3} - 2{\alpha ^2}\Delta {t^2} - 4\alpha \Delta t{e^{ - \alpha \Delta t}}} \right]\\
        q_{12} = \frac{1}{{2{\alpha ^4}}}\left[ {{e^{ - 2\alpha \Delta t}} + 1 - 2{e^{ - \alpha \Delta t}} + 2\alpha \Delta t {e^{ - \alpha \Delta t}} - 2\alpha \Delta t + {\alpha ^2}\Delta {t^2}} \right]\\
        q_{13} = \frac{1}{{2{\alpha ^3}}}\left[ {1 - {e^{ - 2\alpha \Delta t}} - 2\alpha \Delta t{e^{ - \alpha \Delta t}}} \right]\\
        q_{22} = \frac{1}{{2{\alpha ^3}}}\left[ {4{e^{ - \alpha \Delta t}} - 3 - {e^{ - 2\alpha \Delta t}} + 2\alpha \Delta t} \right]\\
        q_{23} = \frac{1}{{2{\alpha ^2}}}\left[ {{e^{ - 2\alpha \Delta t}} + 1 - 2{e^{ - \alpha \Delta t}}} \right]\\
        q_{33} = \frac{1}{{2\alpha }}\left[ {1 - {e^{ - 2\alpha \Delta t}}} \right].
        \end{array}
    \end{equation}
    The currect statistical model parameters $\alpha_i$ is the reciprocal of the maneuvering time constant, and $a_{{\rm{max}},i}$ is the maximum possible acceleration.
    
    \subsubsection{Particle sampling}
    \par The particle sampling method of CSKPF method is almost the same with that of CVPF method, except that the CSKFP method assumes that $p\left( {{\bf{x}}_t^i|{\bf{x}}_{t-1}^i} \right)$ obeys Gaussian distribution with mean value $\mathbf{x}_{t|t-1}$ \emph{but} variance $\mathbf{P}_{t|t-1})$, which means the particles are intensive when the covariance is small and sparse when the covariance is large. In this way, the particle sampling is more efficient.
    
    \subsubsection{State correction}
    \par Similar to the CVPF method, the state estimation $\hat{\mathbf{x}}_t$ and its corresponding observation $\mathbf{y}_t$ can be obtained, but the CSKPF loop does not end here. The state estimation and its covariance is then corrected by observation $\mathbf{y}_t$ using Kalman filter, i.e.
    
    \begin{equation}\label{equ:kcf4}
        \mathbf{x}_{t}=\hat{\mathbf{x}}_t+\mathbf{K}_t\left(\mathbf{y}_t-\mathbf{H}\hat{\mathbf{x}}_t\right),
    \end{equation}
    \begin{equation}\label{equ:kcf5}
        \mathbf{P}_{t}=\mathbf{P}_{t|t-1}-\mathbf{K}_t\mathbf{H}\mathbf{P}_{t|t-1},
    \end{equation}
    where 
    \begin{equation}\label{equ:kcf3}
        \mathbf{K}_{t}=\mathbf{P}_{t|t-1}\mathbf{H}^{\top}\left(\mathbf{H}\mathbf{P}_{t|t-1}\mathbf{H}^{\top}+\mathbf{R}\right)^{-1}
    \end{equation}
    is the Kalman gain matrix, $\mathbf{H}$ is the observation matrix and $\mathbf{R}$ is the observation noise matrix.
    
    \subsubsection{Notes}
    \par It should be noted that the current statistical model assumes that the acceleration obeys the modified Rayleigh distribution whose mean value is the prediction of the current acceleration. Therefore, the CSKPF method requires a period of "warm-up" time to estimate the mean acceleration. In practice, at first $T_h$ moment, a tracker uses both CVPF and CSKPF method to predict and correct the state, but only accept the CVPF output. The mean acceleration is updated at every CSKPF tracker step so that the estimation of acceleration converges to a reasonable value after $T_h$ frames. When $t\geq T_h$, only CSKPF runs.
    \par As can be seen, the CSKPF can not only build the trajectories of objects but also estimate their velocity and acceleration at the same time, which are useful in motion analysis. 
    
    \section{Experiments}\label{sec:test}
    In this section, the CVPF and CSKPF method are compared based on the proposed evaluation index on simulated data and verified on the fruitfly data in the real world.
    
\subsection{Evaluation index} \label{sec:evaluation}
The tracking accuracy index MOTA and the tracking precision index MOTP is widely used to evaluate multiple object tracking (MOT) method\cite{leal-taixeMOTChallenge2015Benchmark2015}. Based on MOTA and MOTP, we propose more intuitively index which customizes for 3D trajectories construction, i.e., tracking integrity, continuity and precision.

\par Suppose that the series $\mathcal{P}^i=\{ \mathbf{p}_{t_i}^i, \cdots, \mathbf{p}_{t}^i, \cdots, \mathbf{p}_{T_i}^i \}$ is the $i$-th true trajectory which starts at $t_i$ and end at $T_i$. The $j$-th tracked trajectory is $\mathcal{Q}^j=\{ \mathbf{q}_{t_i}^j, \cdots, \mathbf{q}_{t}^j, \cdots, \mathbf{q}_{T_i}^j\}$. Note that at some moment $t$, $\mathbf{q}_t^j$ may not exist due to miss tracking. Define the position discrepancy between $\mathcal{P}^i$ and $\mathcal{Q}^j$ at moment $t$ by
\begin{equation}
{d_{ij}}\left( t \right) = \left\{ {\begin{array}{*{20}{c}}
    {\left| {{\bf{p}}_t^i - {\bf{q}}_t^j} \right|}&{\mathrm{if}~{\bf{q}}_t^j{\rm{~ exist}}}\\
    {{d_0}}&{{\rm{otherwize}}}
    \end{array}} \right. , i=1,\cdots, N, j=1,\cdots,M,
\end{equation}
where $d_0$ is the threshold distance. Then, the discrepancy between $\mathcal{P}^i$ and $\mathcal{Q}^j$ among the whole tracking is
\begin{equation}
    d_{ij}=\frac{1}{(T_i-t_i)} \sum_{t=t_i}^{T_i} d_{ij}(t).
    \end{equation}

    \par \textbf{Integrity.}~ For the $i$-th true trajectory, more integral tracking means it can be tracked as long as possible. Therefore, define the tracking integrity as possibility that the true trajectories tracked, i.e., 
    \begin{equation}
        \mathrm{Integrity} = \frac{\sum_{i=1}^N{\sum_{t=t_i}^{T_i}{\delta \left( k_t^i\right)}}}{\sum_{i=1}^N {(T_i-t_i)}}, 
    \end{equation}
    where $\delta \left( \cdot \right)$ is Dirac function.
    
    \par \textbf{Continuity.}~ For the $i$-th true trajectory, more continuous tracking means its matching trajectories are as less as possible. Using IDSW$\left(\mathbf{k}^i\right)$ to count the switch times of matching series $\mathbf{k}^i$, the tracking continuity can be represented by
    \begin{equation}
        \mathrm{Continuity} = 1-\frac{\sum_{i=1}^N{\mathrm{IDSW}\left(\mathbf{k}_i\right)}}{\sum_{i=1}^N {(T_i-t_i)}}.
        \end{equation}
    \par \textbf{Precision.}~ For the $i$-th true trajectory, the tracking precision is defined by the average discrepancy at $k_t^i \neq 0$ between true and tracked trajectories, i.e., 
    \begin{equation}
        \mathrm{Precision} =\frac{ \sum_{i=1}^N { \sum_{t=t_i}^{T_i}{d_{ik_t^i(t)}}}}{\sum_{i=1}^N {\sum_{t=t_i}^{T_i} {\sum_{j=1}^M}}{b_{t,ij}}}.
    \end{equation}
    
\subsection{Simulation experiment}
\subsubsection{Generate true trajectories}
\begin{figure}[!t]
\centering
\begin{minipage}[c]{0.48\textwidth}
\centering
\includegraphics[height=5cm]{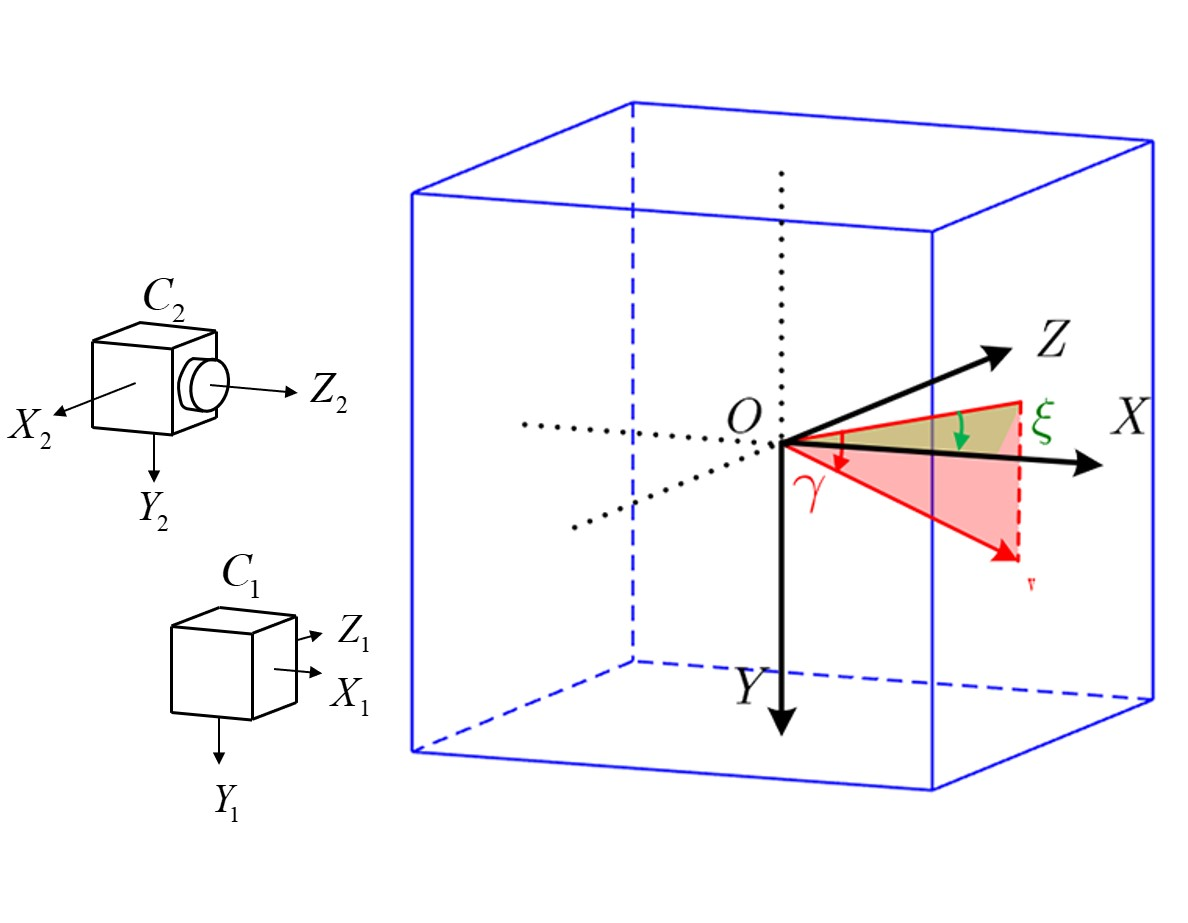}
\end{minipage}
\hspace{0.02\textwidth}
\begin{minipage}[c]{0.48\textwidth}
\centering
\includegraphics[height=5cm]{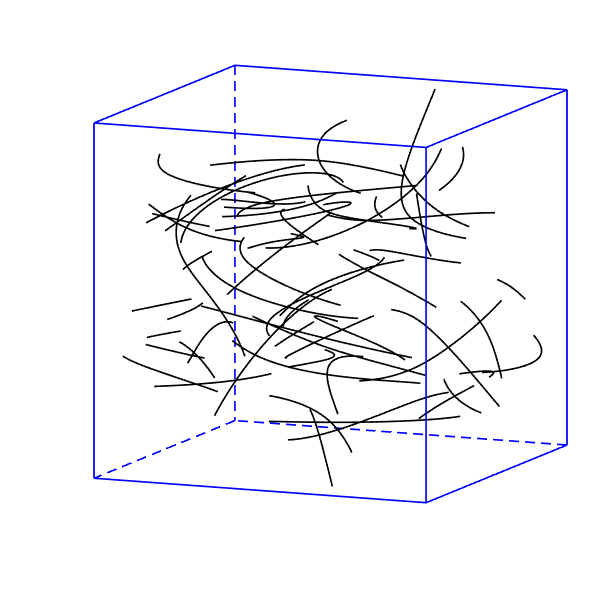}
\end{minipage}\\[3mm]
\begin{minipage}[t]{0.48\textwidth}
\centering
\caption{Coordinate system and defination of heading angle $\xi$ and climbing angle $\gamma$.}
\label{fig:sim_coordinates}
\end{minipage}
\hspace{0.02\textwidth}
\begin{minipage}[t]{0.48\textwidth}
\centering
\caption{Example true value trajectories for $N=60$.}
\label{fig:sim_a}
\end{minipage}
\end{figure}

The simulation data is generated in the coordinate system shown in Figure~\ref{fig:sim_coordinates}. There are $N$ objects whose positions are randomly initialized in the box of $x_0,y_0,z_0 \in [-20,20]$. The $i$-th object's speed is $V_i(t)=6+2\sin\left(\frac{2\pi}{5} t + \alpha_{i,0}\right)$, and its heading angle is $\xi_i(t)=\frac{1}{2} A_{\xi,i} \left(1+\cos\left(\frac{\pi}{10} t + \xi_{i,0}\right)\right)$ and the climbing angle is $\gamma_i(t)=\frac{1}{4}A_{\gamma,i}\cos\left(\frac{\pi}{10} t + \gamma_{i,0}\right)$ at moment $t$, where $\alpha_{i,0}, \xi_{i,0}, \gamma_{i,0},\in [0,2\pi), A_{\xi,i}, A_{\gamma,i} \in [-1,1]$ are randomly initialized. Then, the velocity at moment $t$ is
    \begin{equation}
        \mathbf{v}_i(t)= V_i(t) \begin{bmatrix}
        \cos\gamma_i\cos\xi_i &
        \cos\gamma_i\sin\xi_i &
        \sin\gamma_i
        \end{bmatrix} ^ {\top}
    \end{equation}
The true trajectories are integrated from their velocities.

\subsubsection{Render noised measurements}
\begin{figure}[!t]
\centering
\begin{minipage}[c]{0.48\textwidth}
\centering
\includegraphics[width=6cm]{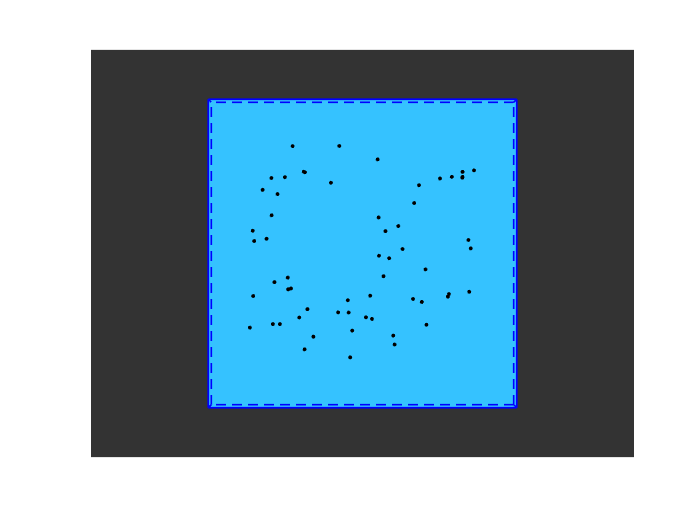}\label{fig:sim_b}
\end{minipage}
\hspace{-5mm}
\begin{minipage}[c]{0.48\textwidth}
\centering
\includegraphics[width=6cm]{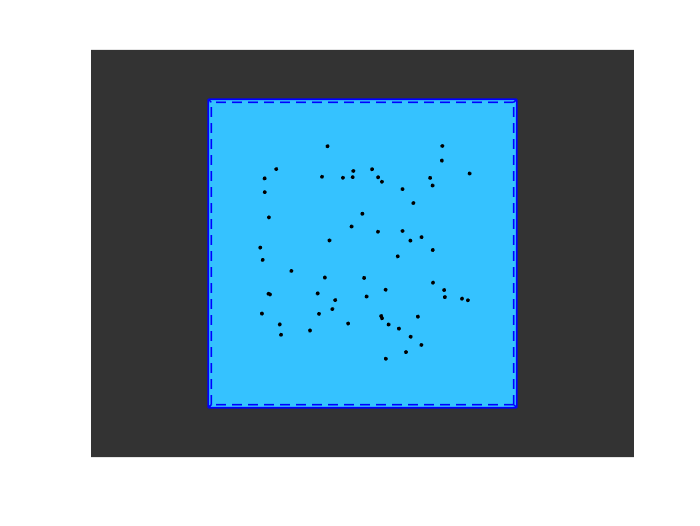}\label{fig:sim_c}
\end{minipage}
\caption{Randered images in (a) view 1 and (b) view 2 for $N=60$ at moment $t=1$. The blue regions are box and the black dots are objects.}\label{fig:sim_render}
\end{figure}

Suppose that the object entity is a ball with a radius of 0.5. As shown in Figure~\ref{fig:sim_coordinates}, two cameras are orthogonally placed, and appropriate camera parameters are selected to ensure that all objects are in the view field of the two cameras. Gaussian noise is added to each object's projection to simulate the measurement error. An example is shown in Figure~\ref{fig:sim_a} and Figure~\ref{fig:sim_render}. 

\subsubsection{Result}
The performance of CVPF and CSKPF are compared when there are $N=$1,20,40,60,80,100, 120, 140, 160 objects and repeats 50 times for each $N$. 
The particle number of the two methods $N_p=100$. The particle covariance $\sigma^2 = 0.09$ in CVPF method and $\alpha=5,a_{\mathrm{max}}= 5$ in CSKPF method. The time step $\Delta t =0.1$s, and the tracking lasts 5 seconds.

\begin{figure}[!t]
\centering
\begin{minipage}[c]{0.33\textwidth}
\centering
\includegraphics[width=4cm]{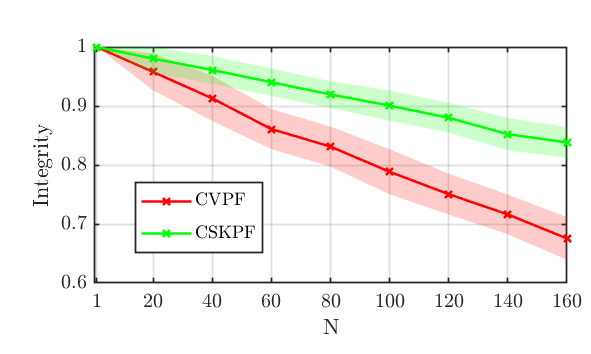}\label{fig:sim_ra}
\end{minipage}
\hspace{-3mm}
\begin{minipage}[c]{0.33\textwidth}
\centering
\includegraphics[width=4cm]{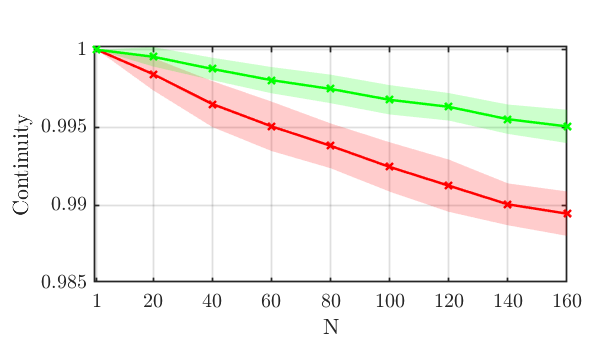}\label{fig:sim_rb}
\end{minipage}
\hspace{-3mm}
\begin{minipage}[c]{0.33\textwidth}
\centering
\includegraphics[width=4cm]{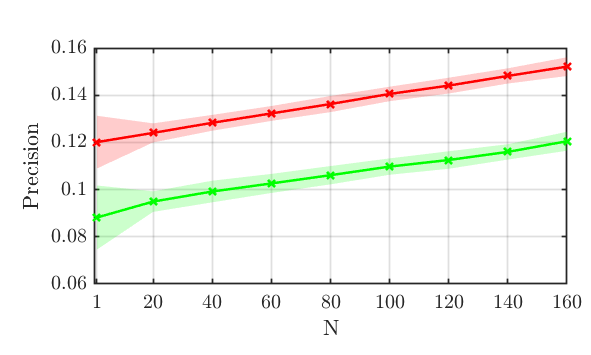}\label{fig:sim_rc}
\end{minipage}
\caption{The tracking (a) integrity, (b) continuity, (c) precision of CVPF and CSKPF method in simulated data. The solid line represents the average index of 50 experiments, and the corresponding shade surrounds one standard deviation.}\label{fig:sim_r}
\end{figure}

\par As shown in Figure~\ref{fig:sim_r}, when there is only one object in the scene, the two methods both successfully build integral and continuous trajectory because there is no temporal-spatial association uncertainty for one object. Nevertheless, the CSKPF method has better precision than the CVPF algorithm. With the increase of objects, the temporal-spatial association uncertainties increases. As a result, the integrity and continuity of the two methods both show a downward trend, and the tracking errors increase. However, the integrity and continuity deterioration of the CSKPF method is more moderate. When there are 160 objects, the CSKPF integrity and continuity can still reach about 0.85 and 0.995, respectively, while that of CVPF has dropped below 0.7 and 0.99. In terms of precision, the tracking errors of the two methods increase linearly, but that of CSKPF is always lower than CVPF. 
\par In summary, the simulation experiments prove that the proposed CSKPF method can improve tracking integrity, continuity, and precision.
    
\subsection{Fruitflies' experiment}
The tracking of the biological swarm is a challenging scenario for the multi-view multi-object tracking method. The CVPF and CSKPF algorithm are further compared on the flight data of fruitflies (\emph{Drosophila melanogaster}) recorded in a laboratory environment.
    
    \subsection{Data acquisition}
    \begin{figure}[htbp]
        \centering
        \includegraphics[width=6cm]{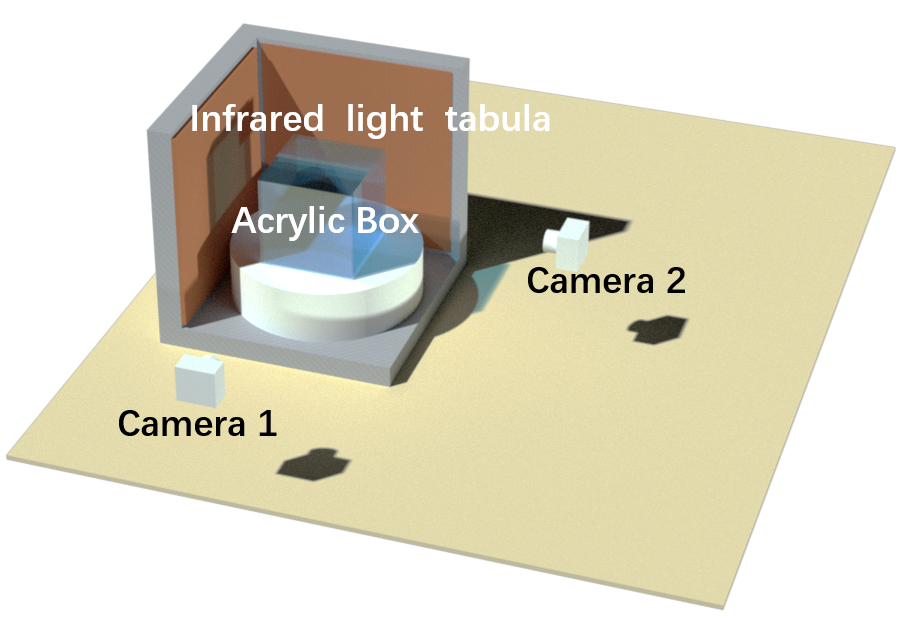}
        \caption{image acquisition system.}\label{fig:scenarios}
    \end{figure}
    
    As shown in the Figure~\ref{fig:scenarios}, the fruitflies fly in a $400 \times 400\times 400$mm box glued by transparent acrylic board. The box is placed between infrared high-speed digital cameras (PCO Dimax HS4 2000x2000@100FPS, Zeiss Biogon T* 35mm f/2 ZM) and the infrared light tabula which are covered with a diffuser to produce soft and flicker-free light. The two calibrated cameras are synchronized and orthogonally placed about 900 mm away from the center of the box.
    
    \par The measurements are expressed as pixel blobs $\chi_{v ,t,k}, v=1,2, t=1,\cdots,T,k=1,\cdots,K$ segmented by Gaussian background modeling method, whcih is the same with \cite{chengNovelMethodTracking2015}. Other detection method \cite{Fu2021} can be used, but it is not discussed here.
    
    \subsubsection{Particle weight}
    Define the geometrical overlap ratio of the particle $\mathbf{x}^i_{t|t-1}$ and the measurement $\chi_{v,t,k_v}$ by
    \begin{equation}
    \eta \left(\mathbf{x}_{t|t-1}^i,\chi_{v,t,k_v} \right) = \frac{|\mathbf{P}^v \Upsilon( \mathbf {x}^i_{t|t-1} )\bigcap \chi_{v,t,k_v}|}{|\chi^i_{v,t,k_v}|},v=1,2
    \end{equation}
    where the $|\cdot|$ operator counts the number of pixels, $\mathbf{P}^v$ is the projection matrix of the camera $v$, $ \Upsilon( \mathbf{x}^i_{t|t- 1} )$ is a ball with the center $\mathbf{x}^i_{t|t-1}$. (The diameter of the ball is set to 3mm because the average body length of fruitfly is about 3 mm). Denote the most likely association at the moment of $t-1$ is $\langle \chi_{1,t-1,k_1^*}, \chi_{2,t-1,k_2^*} \rangle$, the credibility of the $k$-th association ${\langle \chi_{1,t,k_1}, \chi_ {2,t,k_2} \rangle}$ on view $v$ is
    \begin{equation}
    \tau_{v,k}\left( \mathbf{x}_{t|t-1}^i \right) = \eta \left(\mathbf{x}_{t|t-1}^i, \chi_{v,t,k_v} \right)\mathrm{NCC}\left(\chi_{v,t,k_v},\chi_{v,t-1,k_v^*} \right)
    \end{equation}
    where NCC$(\cdot, \cdot)$ is the normalized covariance of the target appearence and the measurement. The credibility of particle $\mathbf{x}_{t|t-1}^i$ on view $v$ is $\tau_v=\mathop{\arg \max} \limits_{k} {\tau_{v,k}}$, and the weight $w_i$ of the particle $\mathbf{x}^i_{t|t-1}$ can be updated by Equation~(\ref{equ:weights}).
    
\subsubsection{Result}
The performance of CVPF and CSKPF are compared on the 100 frames data. The particle number of the two methods $N_p=300$. The particle covariance $\sigma^2 = 4$mm in CVPF method and $\alpha=1,a_{\mathrm{max}}= 0.1$ in CSKPF method. The time step  $\Delta t =0.01$s. 

\begin{figure}[!t]
\centering
\begin{minipage}[c]{0.48\textwidth}
\centering
\includegraphics[width=6cm]{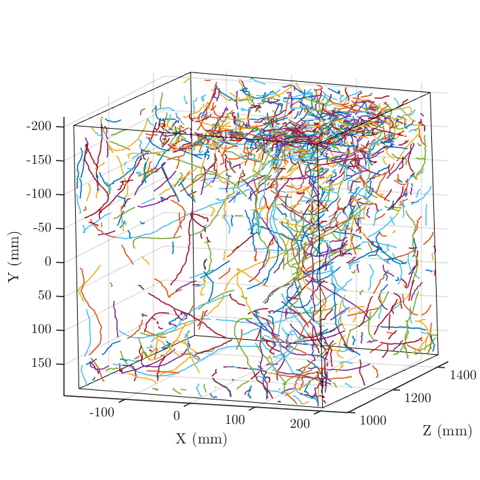}
\end{minipage}
\hspace{0.02\textwidth}
\begin{minipage}[c]{0.48\textwidth}
\centering
\includegraphics[width=6cm]{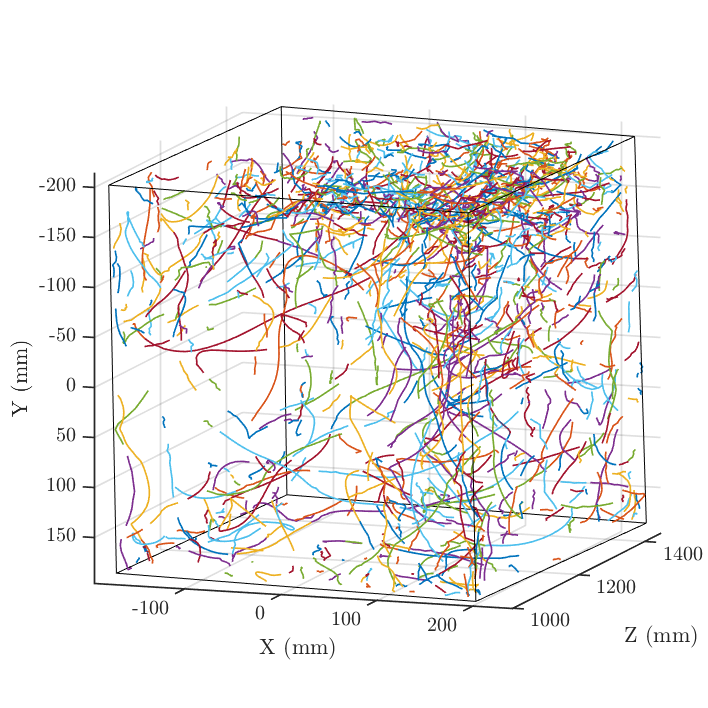}
\end{minipage}
\caption{Trajectories of Fruitfly reconstructed by (a) CVPF and (b) CVKPF method.}\label{fig:real_trajs}
\end{figure}

\par
\begin{figure}[!t]
    \centering
    \begin{minipage}[c]{0.48\textwidth}
        \centering
        \includegraphics[width=6cm]{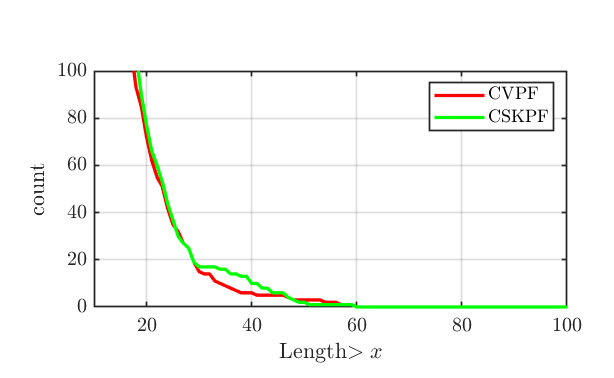}
    \end{minipage}
    \hspace{-5mm}
    \begin{minipage}[c]{0.48\textwidth}
        \centering
        \includegraphics[width=6cm]{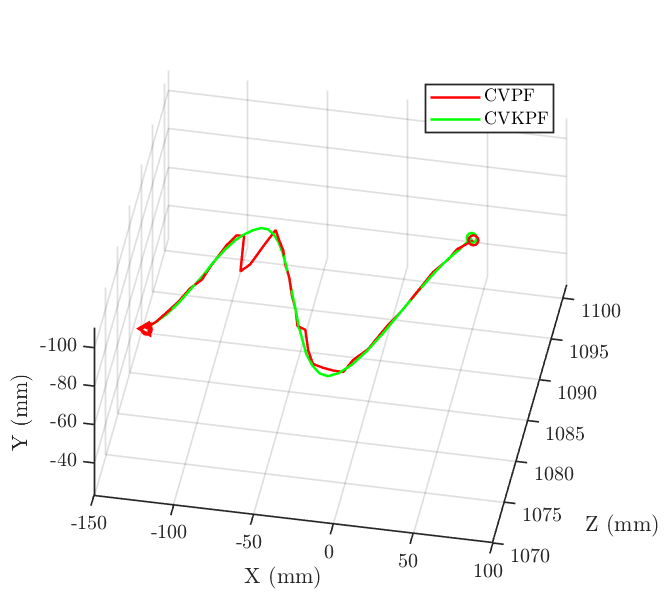}
    \end{minipage}\\[3mm]

    \begin{minipage}[t]{0.48\textwidth}
    \centering
    \caption{Trajectory integrity.}\label{fig:real_integrity}
    \end{minipage}
    \hspace{-5mm}
    \begin{minipage}[t]{0.48\textwidth}
    \centering
    \caption{A typical trajectory. The triangle mark the starting point of the trajectory, and the circles mark the end point of the trajectory.}\label{fig:real_one}
    \end{minipage}
\end{figure}

 As shown in Figure~\ref{fig:real_trajs}, the trajectories of CSKPF are smoother than that of CVPF. Moreover, by counting the trajectories that are longer than $x$ in Figure~\ref{fig:real_integrity}, it can be seen that there are more CSKPF trajectories than CVPF at $30 < x < 50$, which means CSKPF can build a more integral trajectory. It should be noted that at $50 < x < 60$, the integrity of the CVPF is slightly higher because CVPF regards the trajectory that rebounded by the wall of the box as one trajectory while the CSKPF does not accept these kinds of unsmooth trajectories. In practice, only continuous trajectories are concerned. Thus, the integrity of the CSKPF method is better than CVPF. 

\par
\begin{figure}[!t]
\centering
\begin{minipage}[c]{0.33\textwidth}
\centering
\includegraphics[width=4cm]{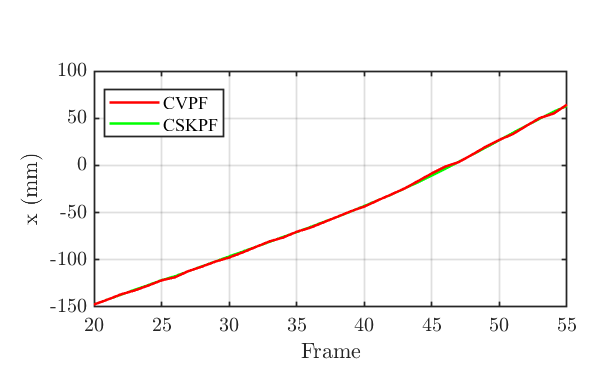}\label{fig:real_x}
\end{minipage}
\hspace{-3mm}
\begin{minipage}[c]{0.33\textwidth}
\centering
\includegraphics[width=4cm]{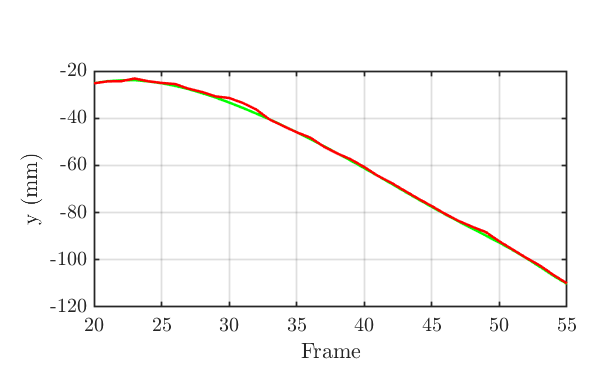}\label{fig:real_y}
\end{minipage}
\hspace{-3mm}
\begin{minipage}[c]{0.33\textwidth}
\centering
\includegraphics[width=4cm]{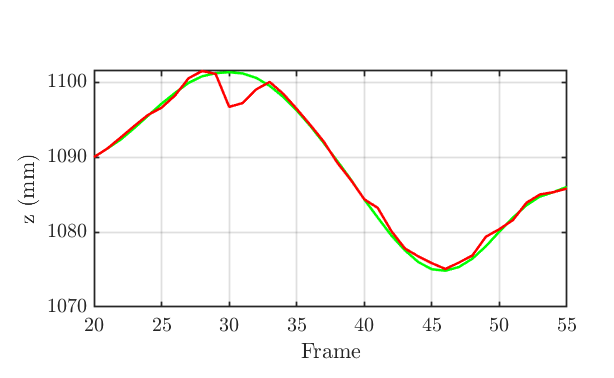}\label{fig:real_z}
\end{minipage}
\caption{Position curve in (a) x, (b) y, (c) z axis of the typical trajectory.}\label{fig:real_one_position}
\end{figure}

\begin{figure}[!t]
\centering
\begin{minipage}[c]{0.33\textwidth}
\centering
\includegraphics[width=4cm]{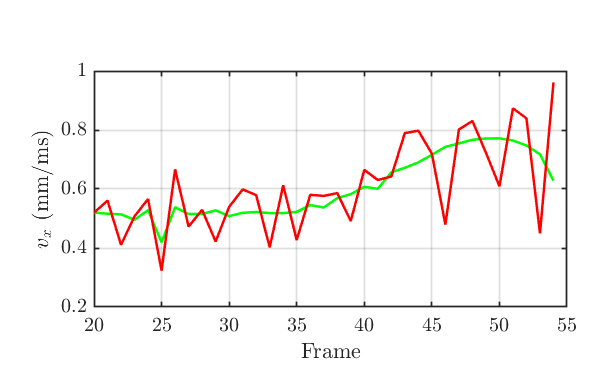}\label{fig:real_vx}
\end{minipage}
\hspace{-3mm}
\begin{minipage}[c]{0.33\textwidth}
\centering
\includegraphics[width=4cm]{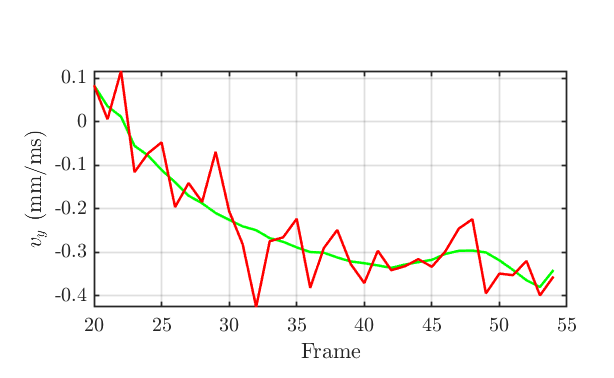}\label{fig:real_vy}
\end{minipage}
\hspace{-3mm}
\begin{minipage}[c]{0.33\textwidth}
\centering
\includegraphics[width=4cm]{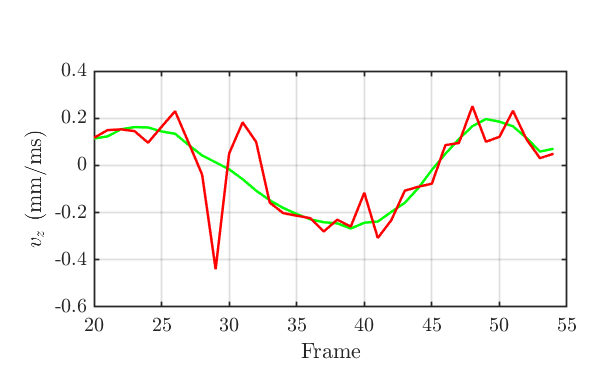}\label{fig:real_vz}
\end{minipage}
\caption{Velocity curve in (a) x, (b) y, (c) z axis of the typical trajectory.}\label{fig:real_one_velocity}
\end{figure}

\begin{figure}[!t]
    \centering
    \begin{minipage}[c]{0.48\textwidth}
        \centering
        \includegraphics[width=5cm]{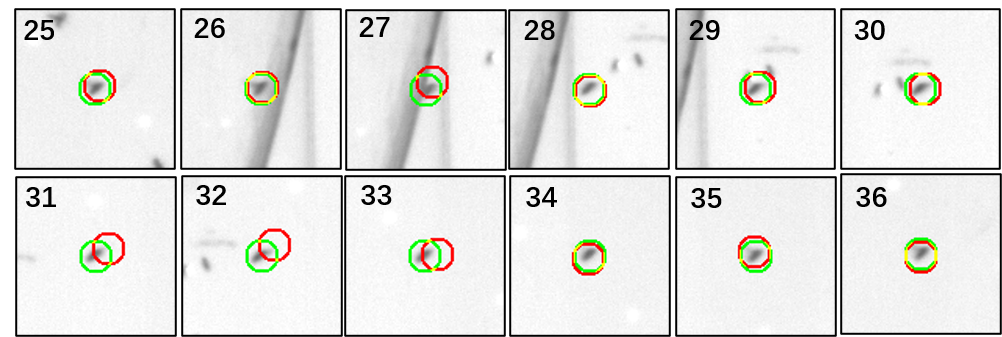}\label{fig:real_reprojection1}
    \end{minipage}
    \hspace{-5mm}
    \begin{minipage}[c]{0.48\textwidth}
        \centering
        \includegraphics[width=5cm]{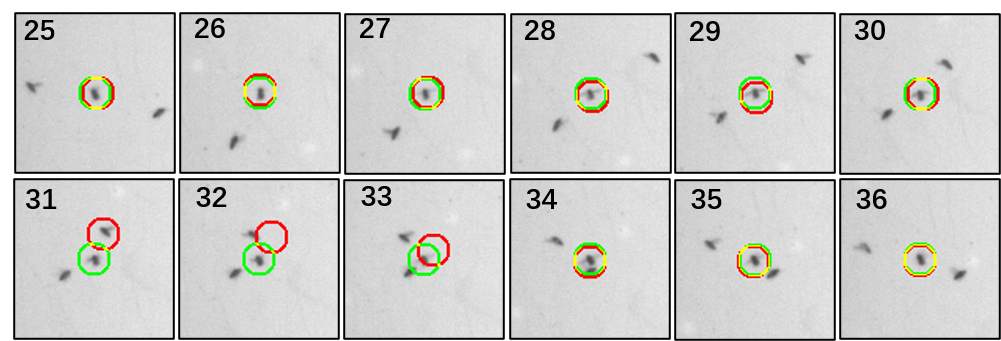}\label{fig:real_reprojection2}
    \end{minipage}
    \caption{Reprojection of the typical trajectory in (a) view 1 and (b) view 2.}\label{fig:reprojection}
\end{figure}

 One typical trajectory is shown in Figure~\ref{fig:real_one} and it position and velocity curve are shown in Figure~\ref{fig:real_one_position} and Figure~\ref{fig:real_one_velocity}. Obviously, the CVPF trajectory is unsmooth and even temporally deviates from the object. The deviation of CVPF occurs around 30 frames according to the position curve. By reporjecting the trajectory between the $25 \sim 36$ frame to each view in Figure~\ref{fig:reprojection}. It is obvious that the deviation is caused by another fruitfly that flew towards the target fruitfly in view 2 at 30 frames. The image of the two fruitflies is too close for CVPF to distinguish while CSKPF successes thanks to better motion prediction which is proved by the smoother and more reasonable velocity prediction. On the other hand, better velocity estimation is important for motion analysis. 
\par In summary, the experiment of fruitfly flight proves that the proposed CSKPF method can more effectively track a large amount of high kinematic objects whose motion model is unknown.
    
\section{Conclusion}\label{sec:conclusion}
Aiming at the multi-view multi-object tracking problem, this paper proposes the CSKPF method which improves the existing CVPF method by introducing the current statistical model to predict the motion, using the state covariance matrix to resample particles and using the Kalman filter to suppress the measurement noise. Both simulation experiments and real-world fruitfly experiments prove the advantages of CSKPF over the baseline CVPF. The proposed method is also suitable for tracking other highly kinematic objects even the kinematic model is unknown. In future research, the missed assumption can be introduced to deal with the trajectory interruption caused by the missed detection, and other fast algorithms to solve the Bayesian inference problem like unscented Kalman filter can be applied to improve the calculation efficiency.

\section*{Acknowledgments}
None

\bibliographystyle{unsrt}  
\bibliography{template}

\end{document}